\documentclass[11pt,a4paper]{article}
\usepackage[hyperref]{acl2020}
\usepackage{times}
\usepackage{latexsym}

\aclfinalcopy 

\usepackage{microtype}

\usepackage{amsfonts}
\usepackage{amssymb}
\usepackage{graphicx}
\usepackage{soul}
\usepackage{comment}
\usepackage{graphicx}
\usepackage{enumitem}
\usepackage{hyperref}
\usepackage{color}
\usepackage{booktabs} 
\usepackage{multirow}

\usepackage{amssymb}

\definecolor{lightblue}{rgb}{.50,.95,1}
\definecolor{tri}{rgb}{.25,.88,.82}
\definecolor{lilac}{rgb}{0.85,0.64,0.85}

\usepackage{subcaption}

\newcommand{\abc}[1]{{\color{magenta} #1}}

\newcommand{\system}{\texttt{Prta}}

\newcommand{\propaganda}{propaganda }

\title{\system: A System to Support the Analysis\\ of Propaganda Techniques in the News}

\author{Giovanni Da San Martino$^{1}$ \quad Shaden Shaar$^{1}$ \quad  Yifan Zhang$^{1}$ \\ 
\textbf{Seunghak Yu}$^{2}$ \quad 
\textbf{Alberto Barr\'{o}n-Cede\~{n}o$^{3}$} \quad \textbf{Preslav Nakov}$^{1}$ \\ $^{1}$ Qatar Computing Research Institute, HBKU, Qatar \\   
$^{2}$ MIT Computer Science and Artificial Intelligence Laboratory,
Cambridge, MA, USA  \\ 
$^{3}$ Universit\`{a} di Bologna, Forl\`{i}, Italy \\
{\tt \{gmartino,sshaar,yzhang,pnakov\}@hbku.edu.qa} 
  \\ {\tt seunghak@csail.mit.edu}\\
  {\tt a.barron@unibo.it}}

\date{}

\begin{document}

\maketitle

\begin{abstract}
    Recent events, such as the 2016 US Presidential Campaign, Brexit and the COVID-19 ``infodemic'', have brought into the spotlight the dangers of online disinformation. 
    There has been a lot of research focusing on fact-checking and disinformation detection. 
    However, little attention has been paid to the specific rhetorical and psychological techniques used to convey propaganda messages. 
    Revealing the use of such techniques can help promote media literacy and critical thinking, and eventually contribute to limiting the impact of ``fake news'' and disinformation campaigns.
    
    \system~(Propaganda Persuasion Techniques Analyzer) allows users to explore the articles crawled on a regular basis by highlighting the spans in which propaganda techniques occur and to compare them on the basis of their use of propaganda techniques. 
    The system further reports statistics about the use of such techniques, overall and over time, 
    or according to filtering criteria specified by the user based on time interval, keywords, and/or political orientation of the media. Moreover, it allows users to analyze any text or URL through a dedicated interface or via an API. The system is available online: \url{https://www.tanbih.org/prta}.
\end{abstract}

\section{Introduction\label{sec:introduction}} 
Brexit and the 2016 US Presidential campaign~\cite{Muller2018}, 
as well as major events such the COVID-19 outbreak~\cite{whocorona}, 
were marked by disinformation campaigns at an unprecedented scale. This has brought the public attention to the problem, which became known under the name ``fake news''.
Even though declared word of the year 2017 by Collins dictionary,\footnote{\abc{\url{https://www.bbc.com/news/uk-41838386}}}
we find that term unhelpful, as it can easily mislead people, and even fact-checking organizations, to only focus on the veracity aspect. 

\noindent At the EU level, a more precise term is preferred, \emph{disinformation}, which refers to information that is both (\emph{i})~false, and (\emph{ii})~intents to harm. 
The often-ignored aspect (\emph{ii}) is the main reasons why disinformation has become an important issue, namely because the news was \emph{weaponized}. 

Another aspect that has been largely ignored is the mechanism through which disinformation is being conveyed: using propaganda techniques. \emph{Propaganda} can be defined as (\emph{i})~trying to influence somebody's opinion, and (\emph{ii})~doing so on purpose~\cite{IJCAI2020:survey}. Note that this definition is orthogonal to that of disinformation: 
Propagandist news can be both true and false, and it can be both harmful and harmless (it could even be good). Here our focus is on the propaganda techniques: on their typology and use in the news. 

Propaganda messages are conveyed via specific rhetorical and psychological techniques, ranging from leveraging on emotions ---such as using \textit{loaded language} \cite[p.~6]{Weston2000}, \emph{flag waving}~\cite{Hobbs2008}, \emph{appeal to authority}~\cite{Goodwin2011}, slogans~\cite{As2015}, and clich\'{e}s~\cite{Hunter2015}--- to using logical fallacies ---such as \textit{straw men}~\cite{Walton1996} (misrepresenting someone's opinion), \textit{red herring}~\cite[p.~78]{Weston2000},\cite{Aper2009} (presenting irrelevant data), \emph{black-and-white fallacy} \cite{Torok2015} (presenting two alternatives as the only possibilities), and \emph{whataboutism} \cite{Richter2017}.

\begin{table*}[t]
\small
\centering
\begin{tabular}{p{\textwidth}}
\toprule
\bf Technique	$\bullet$ Snippet \\
\midrule
\texttt{loaded language}	$\bullet$ \textbf{Outrage} as Donald Trump suggests injecting disinfectant to kill virus.\\
\texttt{name calling, labeling}	$\bullet$ 
WHO: Coronavirus emergency is '\textbf{Public Enemy Number 1}'\\
\texttt{repetition}			$\bullet$ 	I still have a \textbf{dream}. It is a \textbf{dream} deeply rooted in the American \textbf{dream}. I have a \textbf{dream} that one day \ldots\\
\texttt{exaggeration, minimization}	$\bullet$ 
Coronavirus \textbf{'risk to the American people remains very low'}, Trump said.\\
\texttt{doubt}			$\bullet$ \textbf{Can the same be said for the 
Obama Administration}?	\\
\texttt{appeal to fear/prejudice}	$\bullet$ \textbf{A dark, impenetrable 
and “irreversible” winter of persecution of the faithful by their own shepherds 
will fall}.	\\
\texttt{flag-waving}			$\bullet$ Mueller attempts \textbf{to 
stop the will of We the People}!!! It's time to jail Mueller. 	\\
\texttt{causal oversimplification}	$\bullet$ \textbf{If France had not have declared war on Germany then World War II would have never happened.} \\
\texttt{slogans}			$\bullet$ \textbf{``BUILD THE WALL!''} 
Trump tweeted.\\
\texttt{appeal to authority}		$\bullet$ \textbf{Monsignor 
Jean-François Lantheaume, who served as first Counsellor of the Nunciature in 
Washington, confirmed that ``Viganò said the truth. That's all.''}	\\
\texttt{black-and-white fallacy}	$\bullet$ Francis said these words: 
“\textbf{Everyone is guilty for the good he could have done and did not do 
\ldots If we do not oppose evil, we tacitly feed it}. 	\\
\texttt{obfuscation, Intentional vagueness, Confusion}	$\bullet$ \textbf{Women and men are physically and emotionally different. The sexes are not ``equal,'' then, and therefore the law should not pretend that we are!}	\\
\texttt{thought-terminating cliches}	$\bullet$ \textbf{I do not really see 
any problems there.} Marx is the President.	\\
\texttt{whataboutism}		$\bullet$ President Trump ---\textbf{who 
himself avoided national military service} in the 1960's--- keeps beating the 
war drums over North Korea.	\\ 
\texttt{reductio ad hitlerum}	$\bullet$ ``Vichy journalism,'' a term which 
now fits so much of the mainstream media. \textbf{It collaborates in the same 
way that the Vichy government in France collaborated with the Nazis.}	\\
\texttt{red herring}			$\bullet$ \textbf{``You may claim that the death penalty is an ineffective deterrent against crime -- but what about the victims of crime? How do you think surviving family members feel when they see the man who murdered their son kept in prison at their expense? Is it right that they should pay for their son's murderer to be fed and housed?''} \\
\texttt{bandwagon}			$\bullet$ He tweeted, ``\textbf{EU no 
longer considers \#Hamas a terrorist group. Time for US to do same.''} 	\\
\texttt{straw man}			$\bullet$ ``Take it seriously, but with 
a large grain of salt.'' \textbf{Which is just Allen's more nuanced way of 
saying: ``Don't believe it}.''	\\
\bottomrule
\end{tabular}
\caption{Our 18 propaganda techniques with example snippets. The propagandist span appears highlighted. \label{tab:instances}}
\end{table*}

Here, we present \system~---the PRopaganda persuasion Techniques Analyzer. 
\system~makes online readers aware of propaganda by automatically detecting the 
text fragments in which propaganda techniques are being used as well as the type 
of propaganda technique in use. We believe that revealing the use of such 
techniques can help promote media literacy and critical thinking, and eventually 
contribute to limiting the impact of ``fake news'' and disinformation campaigns.

With \system, users can explore the contents of articles about a number of 
topics, crawled from a variety of sources and updated on a regular basis, and to 
compare them on the basis of their use of propaganda techniques. The application 
reports overall statistics about the occurrence of such techniques, as well as 
their usage over time, or according to user-defined filtering criteria such 
as time span, keywords, and/or political orientation of the media. 
Furthermore, the application allows users to input and to analyze any 
text or URL of interest; this is also possible via an API, which allows other 
applications to be built on top of the system.

\system~relies on a supervised multi-granularity gated BERT-based model, which we train on a corpus of news articles annotated at the fragment level with 18 
propaganda techniques, a total of 350K word tokens~\cite{EMNLP2019:propaganda:finegrained}.

Our work is in contrast to previous efforts, where propaganda has been tackled 
primarily at the article level 
\cite{rashkin-EtAl:2017:EMNLP2017,AAAI2019:proppy,BARRONCEDENO20191849}.
It is also different from work in the related field of computational 
argumentation, which deals with some specific logical fallacies related to 
propaganda, such as \textit{ad hominem} fallacy \cite{Habernal2018}.

Consider the game \textit{Argotario}, which educates people to recognize and create 
fallacies such as \textit{ad hominem}, \textit{red herring} and 
\textit{irrelevant authority}, which directly relate to 
propaganda~\cite{Habernal.et.al.2017.EMNLP,Habernal2018b}. Unlike them, we have 
a richer inventory of techniques and we show them in the context of actual news.

The remainder of this paper is organized as follows. Section~\ref{sec:model} introduces the machine learning model at the core of the \system~system. Section~\ref{sec:architecture} sketches the full architecture of \system, with focus on the process of collection and processing of the input articles. Section~\ref{sec:interface} describes the system interface and its functionality, and presents some examples. Section~\ref{sec:conclusions} draws conclusions and discusses possible directions for future work.

\section{Data and Model}
\label{sec:model}

\paragraph{Data} We train our model on a corpus of 350K tokens~\cite{EMNLP2019:propaganda:finegrained,NeurIPS2019:propaganda}, manually annotated by professional annotators with the instances of use of eighteen propaganda techniques. See Table~\ref{tab:instances} for a complete list and examples for each of these techniques.\footnote{Detailed list with definitions and examples is available at\\ \url{http://propaganda.qcri.org/annotations/definitions.html}}

\begin{figure}[t]
\centering
\includegraphics[width=0.95\columnwidth]{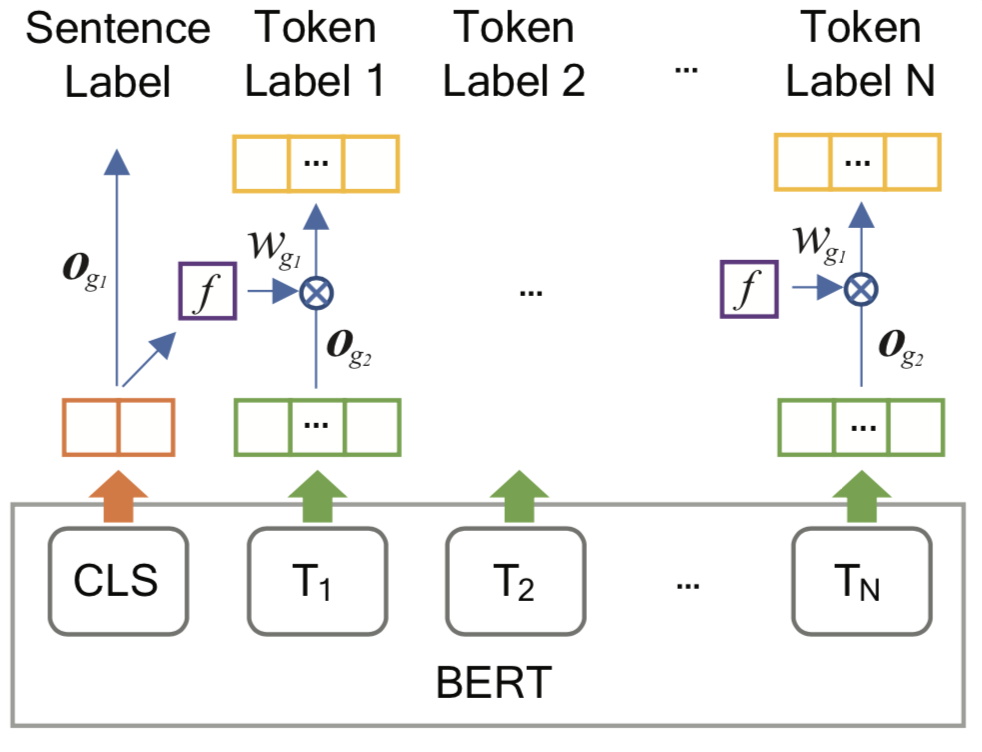}
\caption{The architecture of our model.}
\label{fig:arch}
\end{figure}

\paragraph{Model} Our model is based on multi-task learning with the following two tasks: 
\begin{description}
    \item[FLC] \emph{Fragment-level classification.} Given a sentence, 
    identify all spans of use of propaganda techniques in it and the type of technique.
    \item[SLC] \emph{Sentence-level classification.} Given a sentence, predict whether it contains at least one propaganda technique. 
\end{description}
    
Our model adds on top of BERT~\cite{devlin2018bert} a set of 
layers that combine information from the fragment- and the sentence-level 
annotations to boost the performance of the FLC task on the basis of the SLC 
task. 
The network architecture is shown in Figure~\ref{fig:arch}, and we refer to it 
as a multi-granularity network. 
It features 19 output units for each input token in the FLC task, standing for 
one of the 18 propaganda techniques or ``no technique.''
A complementary output focuses on the SLC task, which is used to generate, through a trainable gate, a weight $w$ that is multiplied by the input of the FLC task.  The gate consists of a projection layer to one dimension and an activation function. 
The effect of this modeling is that if the sentence-level classifier is confident that the sentence does not contain propaganda, i.e.,~$w\sim 0$, then no propaganda technique would be predicted for any of the word tokens in the sentence.

The model we use in \system~outperforms BERT-based baselines on both at the sentence-level ($F_1$ of 60.71 vs. 57.74) and at the fragment-level ($F_1$ of 22.58 vs. 21.39). At the fragment-level, the model outperforms the best solution of a hackathon organized on this data.\footnote{\url{https://www.datasciencesociety.net/events/hack-the-news-datathon-2019}}

\noindent For the \system~system, we applied a softmax operator to turn its output into a 
bounded value in the \mbox{range [0,1]}, which allows us to show a confidence for each prediction.
Further details about the techniques, the model, the data, and the experiments 
can be found in~\cite{EMNLP2019:propaganda:finegrained}.\footnote{The 
corpus and the models are available online at\\  
\url{https://propaganda.qcri.org/fine-grained-propaganda}}

\section{System Architecture}
\label{sec:architecture}

\system~collects news articles from a number of news outlets, discards near-duplicates and finally identifies both specific propaganda techniques and sentences containing propaganda.

We crawl a growing list (now 250) of RSS feeds, Twitter accounts, and websites, and we extract the plain text from the crawled Web pages using the Newspaper3k library\footnote{\url{http://newspaper.readthedocs.io}}. We then perform deduplication based on a combination of URL partial matching and content analysis using a hash function.

Finally, we use the model from Section~\ref{sec:model} to identify sentences with propaganda and instances of use of specific propaganda techniques in the text and their types. 
We further organize the articles into topics; currently, the topics are defined using keyword matching, e.g.,~an article mentioning \textit{COVID-19} or \textit{Brexit} is assigned to a corresponding topic.
By accumulating the techniques identified in multiple articles, \system~can show the volume of propaganda techniques used by each medium ---as well as aggregated over all media for a specific topic--- thus, allowing the user to do comparisons and analysis, as described in the next section.

\begin{figure*}[tbh]
    \centering
    \includegraphics[width=1.0\textwidth]{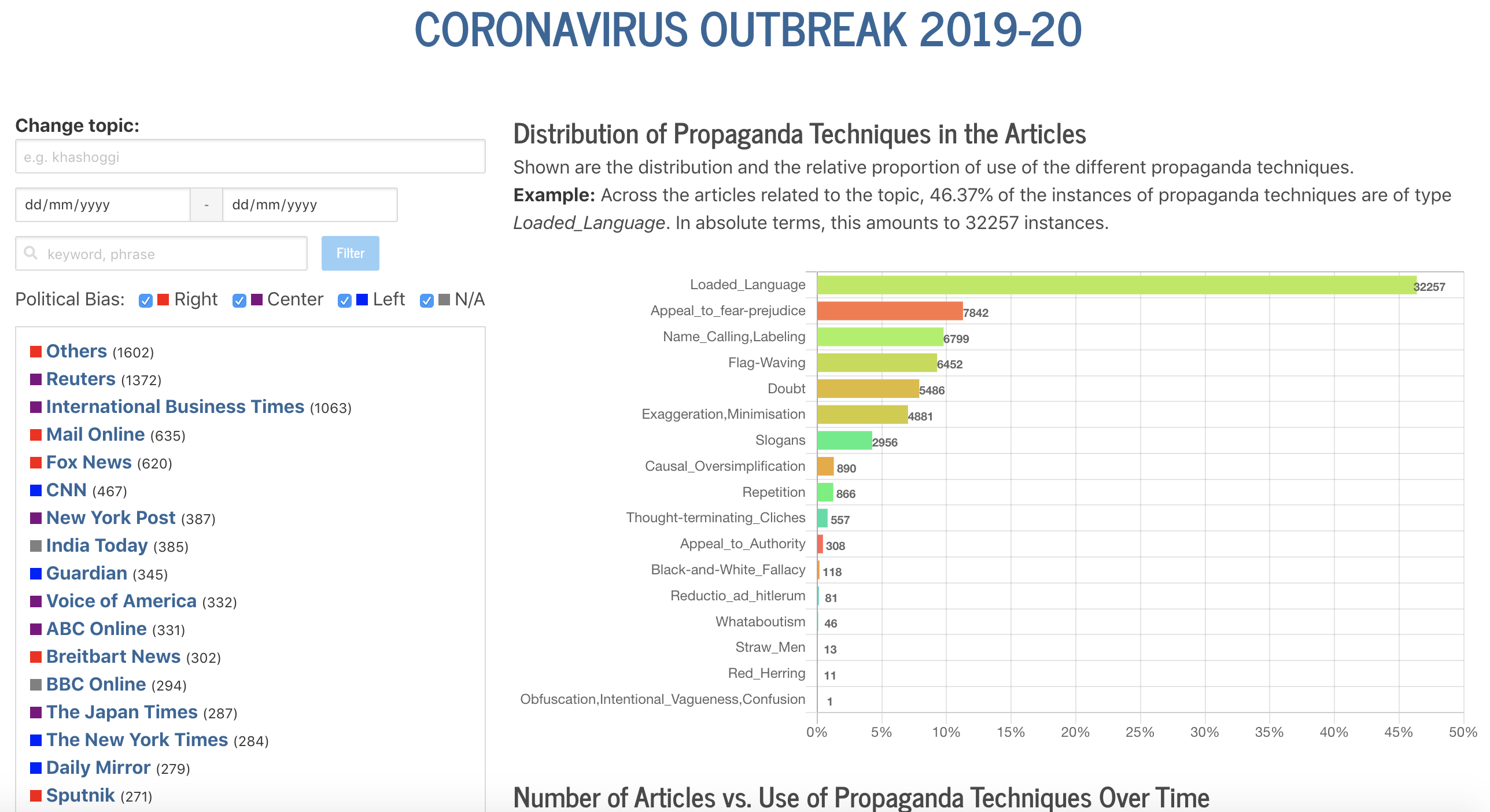}
    \caption{Overall view for a topic. \label{fig:mainpage}}
\end{figure*}

\begin{figure}[t]
    \centering
    \begin{subfigure}[b]{\columnwidth}
    \includegraphics[width=1.0\textwidth]{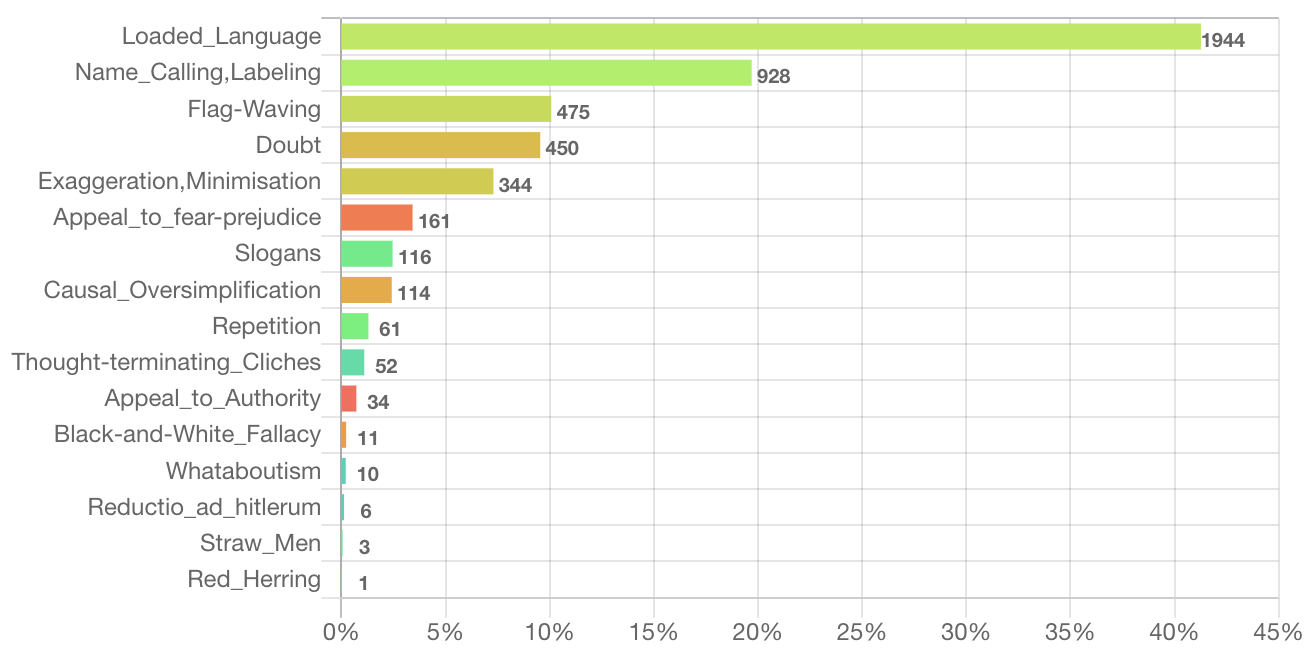}
    \caption{BBC on \emph{Gun Control and Gun Rights}}
        \label{fig:bbc}
    \end{subfigure}
    \begin{subfigure}[b]{\columnwidth}
    \includegraphics[width=1.0\textwidth]{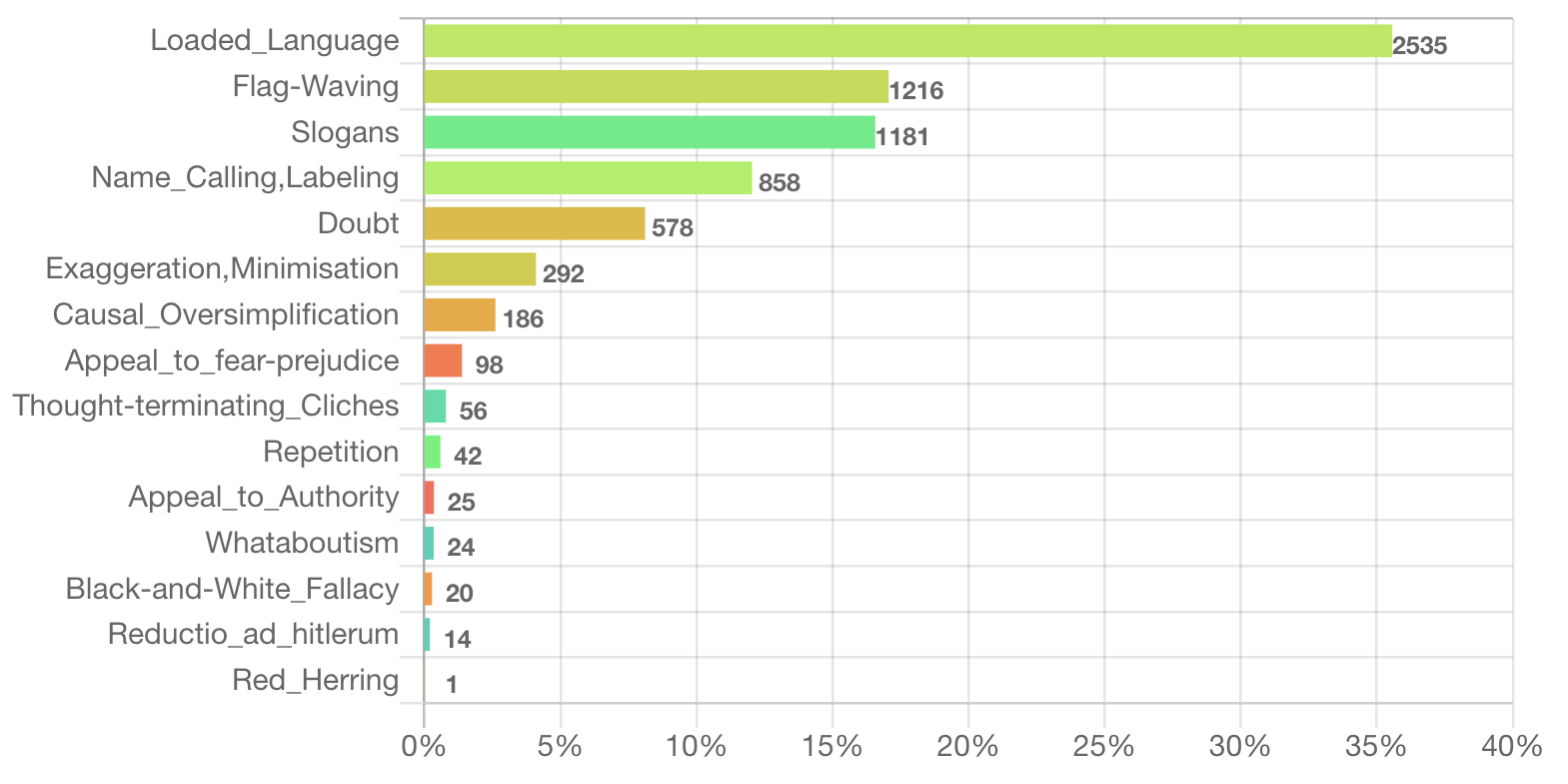}
    \caption{Fox News on \emph{Gun Control and Gun Rights}}
        \label{fig:fox}
    \end{subfigure}
    \begin{subfigure}[b]{\columnwidth}
    \includegraphics[width=1.0\textwidth]{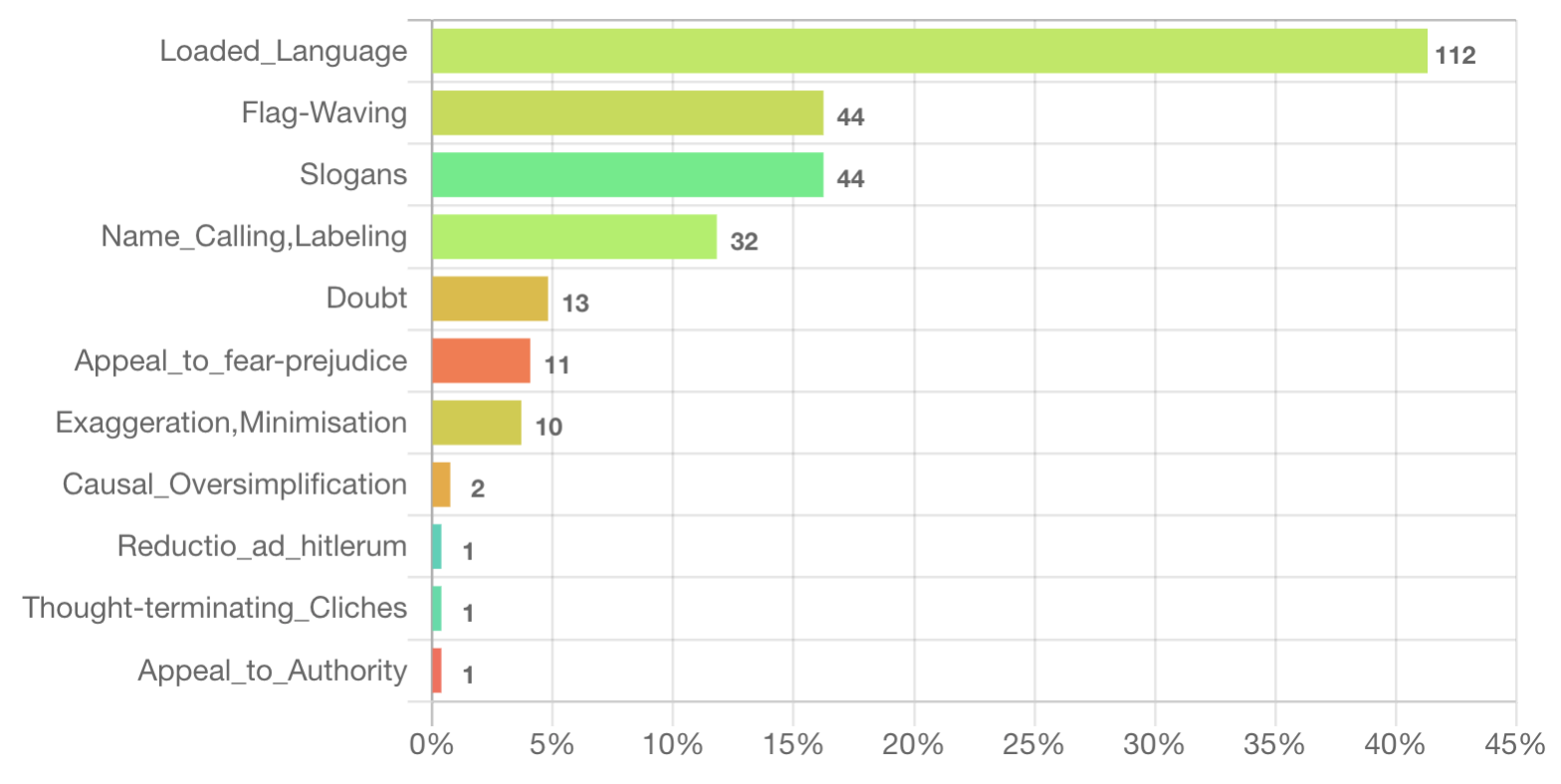}
    \caption{Fox News on \emph{Jamal Khashoggi's Murder}}
        \label{fig:foxKash}
    \end{subfigure}
    \caption{Example of the distribution of the techniques as used by two media 
and on two different topics. Note that the scales 
are different.\label{fig:bbc_fox}}
\end{figure}

\section{Interface\label{sec:interface}}

\system~offers the following functionality.
\smallskip \smallskip

\noindent \textbf{For each crawled news article:}
\begin{enumerate}
    \item\label{it:span} It flags all text spans in which a propaganda technique has been spotted.
    \item\label{it:sent} It flags all sentences containing propaganda. 
\end{enumerate}
\noindent \textbf{For a user-provided text or a URL:}
\begin{enumerate}   \setcounter{enumi}{2}
    \item It flags the same as in~\ref{it:span} and~\ref{it:sent} above.
\end{enumerate}
\noindent \textbf{At the medium and at the topic level:}
\begin{enumerate}   \setcounter{enumi}{3}
\item It displays aggregated statistics about the \propaganda techniques used by all media on a specific topic, and also for individual media, or for media with specific political ideology.
\end{enumerate}

This functionality is implemented in the three interfaces we expose to 
the user: the main topic page, the article page, and the custom article page, which we 
describe in Sections~\ref{sub:main}--\ref{sub:custom}. 
Although points~\ref{it:span} and~\ref{it:sent} above are run offline, they can also be 
invoked for a custom text using our API.\footnote{Link to the API available at \url{https://www.tanbih.org/prta}}

\subsection{Main Topic Page} 
\label{sub:main}

Figure~\ref{fig:mainpage} shows a snapshot of the main page for a given topic: here, the \emph{Coronavirus Outbreak in 2019-20}. We can see on the left panel, a list of the media covering the topic, sorted by number of articles. This allows the user to get a general idea about the degree of coverage of the topic by different media.

The right panel in Figure~\ref{fig:mainpage} shows statistics about the 
articles from the left panel. 
In particular, we can see the global distribution of the propaganda techniques in the articles, both in relative and in absolute terms.
The right panel further shows a graph with the number of articles about 
the topic and the average number of propaganda techniques per article over time. 
Finally, it shows another graph with the relative proportion of propagandistic 
content per article; it is possible to click and to navigate from this graph to 
the target article. The latter two graphs are not shown in 
Figure~\ref{fig:mainpage}, as they could not fit in this paper, but 
the reader is welcome to check them online.

The set of articles on the left panel can be filtered by time interval, by keyword, by political orientation of the media (left/center/right), as well as by any combination thereof.

Clicking on a medium on the left panel expands it, displaying its articles 
ranked on the basis of Eq.~(\ref{eq:scorearticle}). 
Given the output of the multi-granularity network, we compute a simple score to assess the proportion of propaganda techniques in an article or in an individual media source. 
Let $F(x)$ be a set of fragment-level annotations in article $x$, where each 
annotation is a sequence of tokens. 
We compute the propaganda score for $x$ as the ratio between the number 
of tokens covered by some propagandist fragment (regardless of the technique) 
and the total number of tokens in the article:
\begin{equation}
    Q_a(x) = \displaystyle\frac{|\bigcup_{f\in F(x)} f|}{|x|}.
    \label{eq:scorearticle}
\end{equation}

Selecting a medium, or any other filtering criterion, further updates the graph on the center-right panel.
For example, Figures~\ref{fig:bbc} and~\ref{fig:fox} show the distribution of 
the techniques used by the BBC vs. Fox News when covering the topic of \emph{Gun 
Control and Gun Rights}.
We can see that both media use a lot of \texttt{loaded language}, which is the 
most common technique media use in general.
However, the BBC also makes heavy use of \texttt{labeling} and \texttt{doubt}, 
whereas Fox News has a higher preference for \texttt{flag waving} and 
\texttt{slogans}. 

Next, Figure~\ref{fig:foxKash} shows the propaganda techniques used by Fox News 
when covering the Khashoggi's Murder, which has a very similar technique distribution to the plot in Figure~\ref{fig:fox}.

This similarity between the distribution of propaganda techniques in Figures~\ref{fig:fox} and~\ref{fig:foxKash} might be a coincidence, or it could represent a consistent style, regardless of the topic. We leave the exploration of this and other hypotheses to the interested user, which is an easy exercise with the \system~system.

\subsection{Article Page\label{sec:article-analysis}}

\begin{figure*}[tbh]
    \centering
    \includegraphics[width=\textwidth]{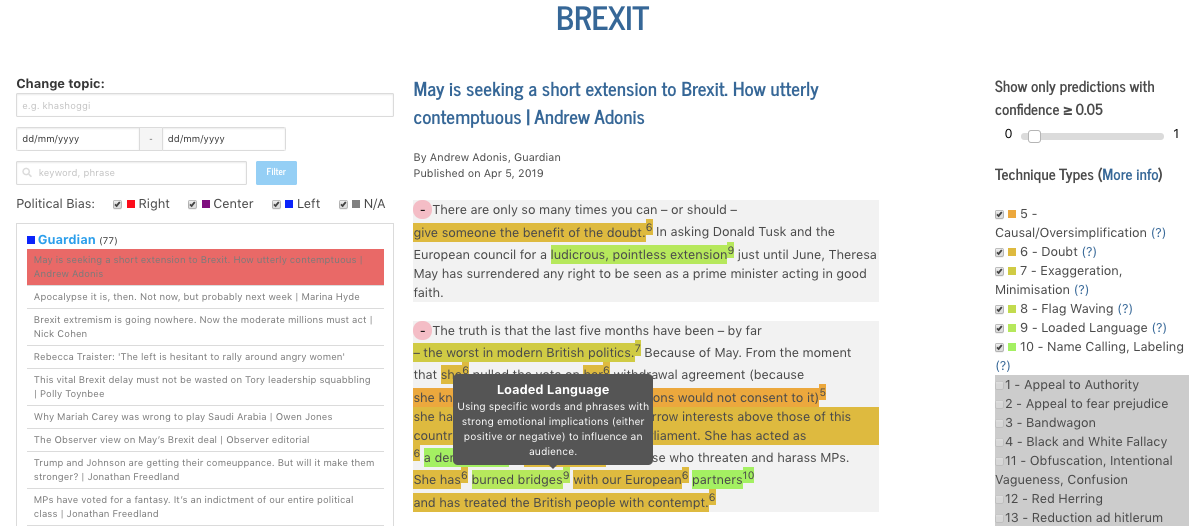}
    \caption{Selecting an article from the left panel, loads it and highlights its propaganda techniques. \label{fig:article}}
\end{figure*}

When the user selects an article title on the left panel (Figure~\ref{fig:mainpage}), its full content will appear on a middle panel with the propaganda fragments highlighted, as shown in Figure~\ref{fig:article}. 
Meanwhile, a right panel will appear, showing the color codes used for each of the techniques found in the article (the techniques that are not present are shown in gray). 

\noindent Moreover, using the slider bar on top of the right panel, the user can set a confidence threshold, and then only those propaganda fragments in the article whose confidence is equal or higher than this set threshold would be highlighted. 
When the user hovers the mouse over a propagandist span, a short 
description of the technique would pop up. If the user wishes to find more information about the propaganda techniques, she can simply click on the corresponding question mark in the right panel.

\subsection{Custom Article Analysis}
\label{sub:custom}

Our interface allows the user to submit her own text for analysis. This allows her to find the techniques used in articles published by media that we do not currently cover or to analyze other kinds of texts. 
Texts can be submitted by copy-pasting in the text box on top, or, alternatively, by using a 
URL. In the latter case, the text box will be automatically filled with the 
content extracted from the URL using the Newspaper3k library (see 
Section~\ref{sec:architecture}), but the user can still edit the content before 
submitting the text for analysis.
The maximum allowed length is the one enforced by the browser. Yet, we recommend to keep texts shorter than 4k in order to avoid blocking the server with too large requests.

\begin{figure*}[ht!]
    \centering
    \includegraphics[width=\textwidth]{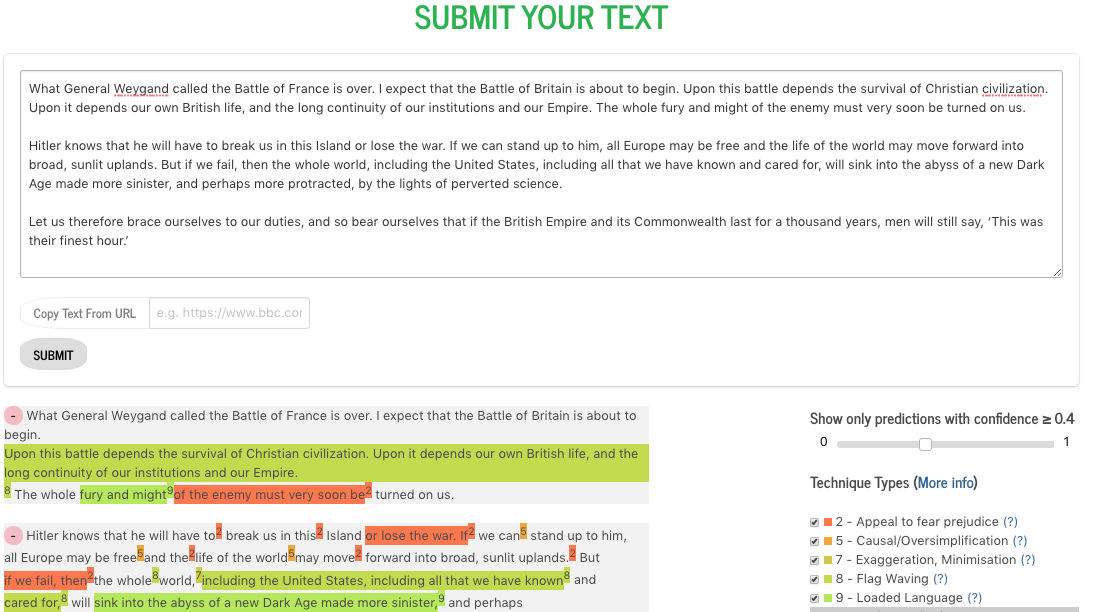}
    \caption{Analysis of a custom text, an excerpt from a speech by W.\ Churchill at the beginning of World War II. The confidence threshold is set to 0.4, and thus fragments for which the confidence is lower are not highlighted. 
    \label{fig:churchill}}
\end{figure*}

Figure~\ref{fig:churchill} shows the analysis for an excerpt of Winston Churchill's speech on May 10, 1940. 
All the techniques found in this speech are highlighted in the same way as described in Section~\ref{sec:article-analysis}. 
Notice that, in this case, we have set the confidence threshold to 0.4 and some of the techniques are consequently not highlighted.
We can see that the system has identified heavy use of propaganda techniques. In particular, we can observe the use of \textit{Flag Waving} and \textit{Appeal to Fear}, which is understandable as the purpose of this speech was to prepare the British population for war.

\section{Conclusion and Future Work} \label{sec:conclusions}

We have presented the \system~system for detecting and highlighting the use of 
propaganda techniques in online news. The system further shows aggregated 
statistics about the use of such techniques in articles filtered according to 
several criteria, including date ranges, media sources, bias of the sources, and 
keyword searches. The system also allows users to analyze their own text or the contents of a URL of interest.

We have made publicly available our data and models, as well as an API to the live system.

We hope that the \system~system would help raise awareness about the use of propaganda techniques in the news, thus promoting media literacy and critical thinking, which are arguably the best long-term answer to ``fake news'' and disinformation.

In future work, we plan to add more media sources, especially from non-English media and regions. We further want to extend the tool to support other propaganda techniques such as \emph{cherry-picking} and \emph{omission}, among others, which would require analysis beyond the text of a single article.

\section*{Acknowledgments}
The \system~system is developed within the Propaganda Analysis
Project\footnote{\url{http://propaganda.qcri.org}}, part of the Tanbih project\footnote{\url{http://tanbih.qcri.org}}. Tanbih aims to limit the effect of ``fake 
news'', propaganda, and media bias by making users aware of what they are reading, thus promoting media literacy and critical thinking. Different organizations collaborate in Tanbih, including the Qatar Computing Research Institute (HBKU) and MIT-CSAIL. 

\bibliography{main}
\bibliographystyle{acl_natbib}

\end{document}